\documentclass[10pt,twocolumn,letterpaper,final]{article}

\usepackage{iccv}
\usepackage{times}
\usepackage{epsfig}

\usepackage{graphicx}
\usepackage{float}
\usepackage{amsmath}
\usepackage{amssymb}
\usepackage{booktabs}
\usepackage{multicol,multirow, color}
\usepackage[pagebackref=true,breaklinks=true,letterpaper=true,colorlinks,bookmarks=false]{hyperref}

\usepackage[capitalize]{cleveref}
\crefname{section}{Sec.}{Secs.}
\Crefname{section}{Section}{Sections}
\Crefname{table}{Table}{Tables}
\crefname{table}{Tab.}{Tabs.}

 \iccvfinalcopy 

\def\iccvPaperID{**} 

\begin{document}

\newcommand{\ourmethod}{\textsf{GATHER}\xspace}
\newcommand{\ourdataset}{ConceptVQA\xspace}
\definecolor{color-q}{RGB}{60,125,195} 
\definecolor{color-v}{RGB}{127,105,175} 
\definecolor{color-n}{RGB}{136,169,192} 
\definecolor{color-a}{RGB}{244,124,100} 
\title{Open-Set Knowledge-Based Visual Question Answering with Inference Paths}

\author{Jingru Gan$^1$, Xinzhe Han$^2$, Shuhui Wang$^{1*}$, Qingming Huang$^2$\\
$^1$ Key Lab of Intell. Info. Process., Inst. of Comput. Tech., CAS\\
$^2$ University of Chinese Academy of Sciences\\
}
\maketitle
\ificcvfinal\thispagestyle{empty}\fi

\begin{abstract}
Given an image and an associated textual question, the purpose of Knowledge-Based Visual Question Answering (KB-VQA) is to provide a correct answer to the question with the aid of external knowledge bases.
Prior KB-VQA models are usually formulated as a retriever-classifier framework, where a pre-trained retriever extracts textual or visual information from knowledge graphs and then makes a prediction among the  candidates.
Despite promising progress, there are two drawbacks with existing models.
Firstly, modeling question-answering as multi-class classification limits the answer space to a preset corpus and lacks the ability of flexible reasoning. 
Secondly, the classifier merely consider ``what is the answer'' without ``how to get the answer'', which cannot ground the answer to explicit reasoning paths.
In this paper, we confront the challenge of \emph{explainable open-set} KB-VQA, where the system is required to answer questions with entities at wild and retain an explainable reasoning path.
To resolve the aforementioned issues, we propose a new retriever-ranker paradigm of KB-VQA, Graph pATH rankER (\ourmethod for brevity).
Specifically, it contains graph constructing, pruning, and path-level ranking, which not only retrieves accurate answers but also provides inference paths that explain the reasoning process.
To comprehensively evaluate our model, we reformulate the benchmark dataset OK-VQA with manually corrected entity-level annotations and release it as \ourdataset.
Extensive experiments on real-world questions demonstrate that our framework is not only able to perform open-set question answering across the whole knowledge base but provide explicit reasoning path.
\end{abstract}

\section{Introduction}
\label{sec:intro}

\begin{figure}[t]
  \centering
  \includegraphics[width=\linewidth]{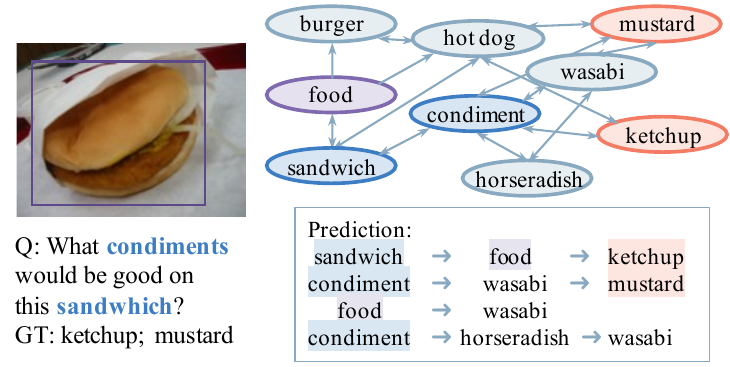}
  \caption{An example of Knowledge-Based Visual Question Answering (KB-VQA). The correct answer is obtained via reasoning on the given question, image, and external knowledge.}
  \label{fig:teaser}
\end{figure}

The ability to reason and to correctly answer questions according to external knowledge are the crucial ability for Visual Question Answering~(VQA) models when challenged with broader usage. 
To boost the reasoning ability, VQA models resort to well-established knowledge graphs~(KGs) such as ConcepetNet~\cite{Liu:2004cn} and DBpedia~\cite{Auer:2007db} which provide abundant external knowledge. 
The resulting Knowledge-Based VQA (KB-VQA) models are capable of incorporating more information of entities and relations within knowledge graphs, exceeding the limit of textual and visual information leveraged from the given image-question pairs, as shown in~\cref{fig:teaser}.
We think explicitly using knowledge bases is a potential solution to aid the reliability of data-driven AI systems, especially for fact-based problems.

To date, the retriever-classifier framework is still the dominant choice for KB-VQA. 
It consists of a hand-crafted knowledge retriever and a classifier~\cite{Shevchenko:2021ro, Wu:2021ma, Ding:2022vt} or extraction text reader~\cite{Luo:2021up}. 
With the successful attempts of transferring KB-VQA into textual domain, many studies take their focus on exploring multi-modal information with pre-trained vision-language models~\cite{li2019visualbert,Lu:2019vilbert,Tan:2019lx,wang2021simvlm,zeng2022socratic} or advocate reasoning with Graph Neural Networks (GNNs)~\cite{Lin:2019vf,Wang:2022bo, Feng:2020tl} over multimodal knowledge due to the structured nature of knowledge bases.
Although modeling KB-VQA as classification makes the overall framework feasible for end-to-end learning and achieve impressive performance on benchmarks~\cite{Shevchenko:2021ro, Wu:2021ma, Ding:2022vt}, the answer space is narrowed down to a small pre-defined range of candidates and the reasoning process is kept in a black box. 
This is completely against the significance of KB-VQA that is to extend the applicability of VQA to \emph{reasoning} over \emph{large-scale} external knowledge bases. 
The labeled answers in training data is far less from the number of entities in the knowledge graphs ({\it e.g.}, 5,020 in OK-VQA~\cite{Marino:2019ug} vs. 516,782 in ConceptNet~\cite{Liu:2004cn}).
To extend the generalizability, some other works explore open-ended VQA with a generative models. 
They stress the feasibility of pre-trained LLMs~\cite{Yang:2022ae,Jin:2022AGP,Gui:2022kat} where the extremely large-scale implicit knowledge accumulated allows the model to generate an open-ended answer. 
But these methods suffer from poor reliability and expansibility since these kind of ``zero-shot'' learning relies on the large-scale pre-train corpus rather than the explicit structured external knowledge.

We argue that the capability of KB-VQA should not be limited by the number of annotated QA pairs but the scale and the structure of the external knowledge. 
Back to the motivation of KB-VQA at the very beginning, we try to investigate two following questions in this paper, the first is how a model pick the correct answer from the whole massive KB rather than just the annotated labels, and the second is
how the VQA make use of the structure of KGs and comes to an interpretable reasoning process. 
To this end, we propose a new retriever-ranker paradigm, Graph pATH rankER (\ourmethod), that takes a step ahead on both challenges. 
To handle billions of entities and relations in KG, a cascade retriever is first proposed to filter the answer candidates from coarse-to-fine. 
Concretely, we first retrieves all image-related information from the vast knowledge base and construct a scheme graph that inherits the structure of both image scene graph and the KG. 
Then, the scheme graph is further pruned to accentuate on question-related information. 
On top of the distilled subgraph, we further introduce a new path-level ranking method for an explainable reasoning path. Different from traditional retriever-classification frameworks, \ourmethod exceeds the pre-defined answering space and tries to provide reasoning paths apart from predicted answers.

To evaluate the performance of open-set KB-VQA, we present a new benchmark ConceptQA, a re-labeled subset from OK-VQA~\cite{Marino:2019ug}. 
All answers in \ourdataset are covered by the entities in ConceptNet while 160 answers in \texttt{open test} split have no overlap with the answer annotations in the training set. 
The classification-based methods will completely fail on this split.
We use the metric of recall of ground truth entity at top-$k$ retrieval for evaluation. The experiments demonstrate that the extention of retrieving answer from the wild is feasible and of great potential.
We find that the schema graph construction and graph pruning process show promising hit rate of answer out of the entire KG. 
Moreover, our path inference module provides an interpretable reasoning result which shows that explicit inference on a knowledge graph may be a potential complementary for data-driven models in the future. 

Our contribution is three-fold.




\begin{itemize}
	\item We advocate that open-set answers and the ability of reasoning are the keys of taking real use of the KGs. Accordingly, we present a new setting, open-set KB-VQA, where the answer space is no longer the labeled answers but the concepts across the knowledge graph.
	
	\item We present a new benchmark, ConceptVQA, in which any question is able to be answered with the entities in ConceptNet but the some answers in test split have never appeared during training. It can evaluate whether a model can generalize to open-set answers on the whole knowledge graph.
	\item We propose a retriever-ranker KB-VQA framework \ourmethod that retrieves from coarse knowledge and reasons on clear paths. 
	It is the first attempt to provide explicit reasoning paths and open-set answers in KB-VQA. 
	Experiments on ConceptQA demonstrate that extending the answer space to the whole knowledge graph is feasible and worths more in-depth research.

\end{itemize}

\section{Related Work}

\subsection{A General Solution to KB-VQA}
Researches on KB-VQA have all adopted a two-step retriever-answerer framework, {\it i.e.}, a knowledge retriever first traverses through outside KBs for question or image related facts, and then the knowledge together with given queries are fed into the answerer to predict an answer. 

The knowledge retriever usually takes hand-crafted rules to function, and is not trained end-to-end with the answerer, the structure of which varies with the form of KBs as well as the expected knowledge output.
For text-form knowledge, a dense passage retrieval model (DPR)~\cite{Karpukhin:2020dpr} that are better at retrieving relevant knowledge snippets than TF-IDF and BM25 has prompted many researches to use pre-trained dual-encoder on query and passage~\cite{Luo:2021up, Gao:2022gw}.
On the other hand, the visual knowledge can be obtained by employing visual models on the input image, representative models of which include objective segmentation models~\cite{He:2017mrcnn}, image captioning models~\cite{Luo:2021up}, and Google Image~\cite{Wu:2021ma}.

Regarding the framework of the answerer, previous researches can be grouped into two broad types, i.e., answer generator and answer classifier.
A pre-trained generator, mainly adapted from the QA research domain, generates answers based on plain-text transformed from given image, question, and knowledge~\cite{Gao:2022gw}.
Recent works PICa~\cite{Yang:2021jb} and KAT~\cite{Gui:2022kat} that use GPT-3 as internal knowledge sources prove the efficacy of pre-trained generative language models on VQA tasks.
Nonetheless, they rely on large-scale pre-trained corpus and thus are computational expensive.
Therefore, more KB-VQA approaches build their answerers as classifiers.
An answer classifier predicts the final answer by classifying among a pre-defined answer space, which is usually the union of all ground-truth answers appeared in all data splits. 
Along this line, most researches investigating multi-modal reasoning utilize vision-language models to incorporate visual and linguistic information in the queries and knowledge~\cite{Shevchenko:2021ro, Wu:2021ma, Ding:2022vt}.
Several studies of QA~\cite{Lin:2019vf, Feng:2020tl, Yasunaga:2021ug, Wang:2022wi}  and VQA problems~\cite{Narasimhan:2018ot, Garderes:2020tc, Marino:2021tn, Wang:2022bo} employ GNNs to enhance model explainability, considering the natural graph structure of KGs, \eg, scene graph, ConceptNet~\cite{Liu:2004cn} and DBpedia~\cite{Auer:2007db}.
Such frameworks usually follow a two-step paradigm of \textit{schema graph grounding} and \textit{graph modeling for inference} ~\cite{Lin:2019vf, Wang:2022wi}.
In comparison, our \ourmethod that answers open-set KB-VQA with KG entities differs from retriever-classifiers or retriever-generator on the open answer space and the KG-based inference path.

\subsection{Knowledge-Based Inference}
Researchers propose different solutions in the pursuit of explainable VQA.
GNN-based models such as QA-GNN~\cite{Yasunaga:2021ug}, VQA-GNN~\cite{Wang:2022bo}, and GSC~\cite{Wang:2022wi} reason an answer by propagation and aggregation on a question-image related graph. 
In their models, the inference process can be interpreted by decoding attention weights of graph parameters.
Recent work KVQAmeta~\cite{GarciaOlano:2022ia} improves the model explainability by knowledge-grounded entity linking.
MuKEA~\cite{Ding:2022vt} performs inference on retrieved multi-modal triplets and models the answer prediction as a knowledge graph completion problem, which explicitly highlights reasoning paths.
However, all the above models are confined with a preset close set containing answer candidates, which cannot fully explore the whole knowledge bases and thus lack flexibility.

\subsection{Open-Set VQA}
KB-VQA extends the vanilla VQA in the case of complicated questions that need to query information beyond the scope of the given question and image pair.
However, many existing KB-VQA models need to have a pre-defined answer candidate set which limits their flexibility in fully exploring the external knowledge base.
In pursuit of open-ended QA, extractive reader is a representative model that extracts answers from the retrieved knowledge pieces~\cite{Luo:2021up} by predicting the start and end positions of an answer span in the retrieved knowledge passages.
However, this approach is not directly applicable to VQA due to the lack of annotations on the answer span.
Another way of realizing open-ended VQA is to generate the answer from scratch.
Some strong pre-trained learners \cite{Tsimpoukelli:2021mf, Yang:2022ae,Jin:2022AGP,Gui:2022kat} are proven capable of generating plausible answers, the success of which are concluded on unprecedented amount of implicit knowledge stored in large-scale language models such as GPT-3\cite{Brown:2020gpt3}, the explicit knowledge retrieved respectively for each question and a joint reasoning network.

Regarding the answer space, extractive readers predict start and end position of an answer span in retrieved knowledge passages.
An answer generator predicts word distribution at each position over the vocabulary space.
While our model predicts answer among knowledge entities retrieved from the entire knowledge graph.
However, a common concern is that both approaches rely heavily on a high recall of ground-truth-contained knowledge pieces. 
Model performances are sensitive to the quality and volume of knowledge retrieved~\cite{Luo:2021up}, \eg, 100 knowledge passages~\cite{Gao:2022gw} are prerequisites for an extractive reader to obtain a sufficient knowledge hit rate. The number of passages is so large that would damp the reader accuracy. 
This bottleneck is caused by the lack of knowledge ground-truth annotations to each question. 
Our reformulated \ourdataset of OK-VQA~\cite{Marino:2019ug} dataset, however, provides a solution to supervised knowledge retrieval.





\begin{figure*}
	\centering
	\includegraphics[width=\linewidth]{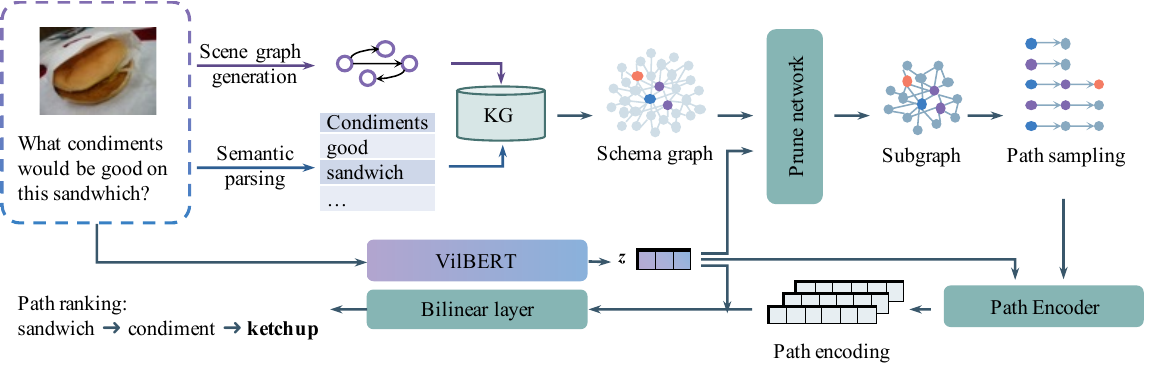}
	\caption{The general framework of our \ourmethod. Our retriever-ranker framework are constructed with schema graph constructing, graph pruning and path inference on subgraph. The origin of each node in the schema graph and subgraph is indicated in color, which are \textcolor{color-q}{question}, \textcolor{color-v}{image} and \textcolor{color-n}{neighborhood}, as well as the \textcolor{color-a}{ground truth} answer.}
	\label{fig:gel_framework}
\end{figure*}

\section{The Proposed \ourmethod Approach}
Given an image $I$, its related textual question $Q$, and a knowledge graph $\mathcal{KG} = (\mathcal{V}, \mathcal{E})$, where the vertex set $\mathcal{V}$ denotes the entities and $\mathcal{E} \subseteq \mathcal{V} \times \mathcal{R} \times \mathcal{V}$ denotes the edge set, the problem of open-set KB-VQA aims to retrieve answers to $Q$ along with reasoning paths from the entities in KGs.
In practice, we use ConceptNet as the external knowledge graph, where $\mathcal{KG}$ contains 516,782 entities and 42 different edge types and their reversions. 
Open-set KB-VQA aims to retrieve answers and reasoning paths from the entities in KGs. 
Our proposed \ourmethod consists of three steps: schema graph construction, subgraph pruning, and reasoning path ranking.
The overall pipeline is shown in \cref{fig:gel_framework}.
\subsection{Schema Graph Construction}
\label{subsec:sgc}



In the schema graph construction phase, we perform node extraction of the textual and visual information respectively, and then bridge and expand the schema graph to an ideal size with neighbor nodes, see \cref{fig:teaser}.
Our knowledge retriever is somehow similar to some of the retriever-classifier methods that convert the questions and images to graphs, reserving the structure information of KGs\cite{Wang:2022bo, Lin:2019vf, Feng:2020tl, Yasunaga:2021ug}. 
A retrieved schema graph is denoted as $\mathcal{G}$.
Nodes of the schema graph are labeled into three types $\mathcal{V} = (\mathcal{V}_Q, \mathcal{V}_V, \mathcal{V}_N)$, each indicating the source of nodes from question, image, and KG, also colored in \textcolor{color-q}{blue}, \textcolor{color-v}{purple} and \textcolor{color-n}{grey} throughout all the figures in this paper.

\textbf{Textual Graph Construction.}
We extract key information from a given question by chunking and part-of-speech tagging.
We keep only the nouns, verbs, adjectives, adverbs as well as phrases.
10 most commonly queried words are discarded, including `which', `picture' and so on.
The extracted question chunks are then matched with ConceptNet entities. The matched entities are added to the schema graph, and then become the question nodes $\mathcal{V}_Q$.

\textbf{Image Graph Construction.} 
Following previous work\cite{Wang:2022bo}, the image information is represented as a scene graph via the pre-trained scene graph generator~\cite{Tang:2020sg}. 
The scene graph is further reduced to the top 30 bounding-box labels and top 20 triplets under the restriction that at least one of the nodes is recruited by excluding the predicate like `has' and `of'.
This is to avoid noisy triplets describing factual knowledge that are normally unrelated to questions, \eg, `(human, has, leg)'.
Then we convert these nodes to entities and relations of ConceptNet according to a matching table.
Nodes or relations with matching form or synonyms will be added to schema graph. 
Each edge are labeled with edge weight which is the confidence score of the detected triplet.

\textbf{External Knowledge Searching.}
Based on retrieved textual and visual nodes, we search for one-hop and two-hop neighbors from ConceptNet, which should intuitively  cover most of the expected answers.
The number of retrieved neighbor nodes could be up to a few thousands, on which we perform sorting sequentially by summed edge weights given in ConceptNet, pre-defined relation order, number of related nodes and number of $\mathcal{V}_Q$ nodes in those related nodes.
Top 500 one-hop nodes with their relations are chosen to build the schema graph.
Similar procedure are undertaken for two-hop neighbors, filling a 1,000-node schema graph. All neighbor nodes are denoted as  $\mathcal{V}_N$.


Each entity in ConceptNet is considered a potential answer to any of the questions, which is the ground of our open-ended KB-VQA.
By constructing a schema graph, we traverse all entities and keep only the ones closely related to the given question and image, which allows maximum degree of open-ended answer while maintains computationally feasible.

\subsection{Graph Pruning}
\label{subsec:prune}

The extracted schema graph is still too prohibitive for a path-level inference.
Therefore, we perform an extra pruning process on the obtained schema graph, which further reduces the number of nodes and avoids high computational burden according to the given vision and language information.
Similar to \cite{Wu:2021ma,Ding:2022vt}, we use VilBERT~\cite{Lu:2019vilbert} to capture visual and textual context of the given question $Q$ and image $I$. It produces visual feature $v\in \mathbb{R}^{d}$ and textual feature $t \in \mathbb{R}^{d}$ and cross-attended multi-modal context feature $z \in \mathbb{R}^{d} $.
\begin{equation}\label{vilbert}
z, v, t = \texttt{VilBERT}(Q,I)
\end{equation}
We take the pre-trained Trans-E embedding of each entity as node entity embedding $e_i \in \mathbb{R}^{D},D=300$.
Then we use a one-hot 4-digit code $u_i \in \{0,1\}^{|\Gamma|}$ with activation as the node type embedding, similar to that of VQAGNN~\cite{Wang:2022bo}.
To strengthen the textual information encoded in query text and candidate answer word, we build a PTM encoding layer~\cite{Jiang:2022jj} out of TinyBERT to encode textual feature of the concatenated question and candidate answer $p_i \in \mathbb{R}^d$. 
The node-level representation is represented as
\begin{equation}\label{node_enc}
h_{i}^{(n)} = f_n([z||e_i||p_i||u_i])
\end{equation} 
where $f_n:\mathbb{R}^{d+D+d+|\Gamma|} \rightarrow \mathbb{R}^d $ is the MLP layers and $||$ operation indicates concatenation among vectors.
The node-question relevance scores between the multimodal context representation $z$ and embedding of each node $h_{i}^{(n)}$ with cosine similarity is calculated as

\begin{equation}
s_{cos} = cos(z, h_{i}^{(n)}).
\end{equation} 

The pruning network is optimized using a triplet loss function, where the anchor in every triplet is the multi-modal context embedding $z$ and the positive term is the ground-truth node.
\begin{equation}
\mathcal{L}_{prune} = triplet(z, h_{n_{gt}}, h_{n_j}).
\end{equation} 

With the node relevant score, we can prune the schema graph $\mathcal{G}$ into an informative and compact subgraph $\mathcal{G}_{sub}$.
To maintain the connectivity of the pruned subgraph and keep the key nodes after pruning, we perform a breadth-first search (bfs) form all the key nodes $n_i \in \{\mathcal{V}_Q \cup \mathcal{V}_V\}$ originated from image and question. 
Then each node will be assigned a bfs score $s_{bfs}$ indicating its distance from these key nodes. 
We use a weighted sum of the two scores to decide which node to be pruned.
\begin{equation}\label{l_prune}
s_{prune} = \theta_{p} s_{bfs} + (1-\theta_p)s_{cos}
\end{equation} 

For a subgraph of size $|\mathcal{V}_{sub}|$, the top $|\mathcal{V}_{sub}|$ nodes with highest scores are maintained.
The graph pruning process can reduce the size of schema graph while leaving few isolated nodes.

\subsection{Inference Path Ranking}
After the schema graph construction and graph pruning, for each question we narrow down the candidate answer to $|\mathcal{V}_{sub}|$, a limited number of entities.
It provides a solid foundation for explainable path-level inference. 

We assume that the inference path starts from question/image nodes and sample $N$ paths within $k$ steps from the key nodes of type $\{Q,V\}$.
The sequential path is encoded as a unified embedding:
\begin{equation}\label{p_encode}
h_{j}^{(t)} = f_t(p_n)
\end{equation} 
where $f_t:\mathbb{R}^{k\times d} \rightarrow \mathbb{R}^{1 \times d} $ is a two-layered MLP.
The final path embedding is calculated from the concatenation with visual and textual context features:
\begin{equation}
h_{j}^{(p)} = f_p([t||v||h_{j}^{(t)}]),
\end{equation}
where $f_p:\mathbb{R}^{d+d+d} \rightarrow \mathbb{R}^d $ is a two-layered MLP.

To optimize the path selection process, we regards paths ended with the grounded-truth answers as the positive path $l_{gt}$, which is a kind of weak supervision for path selection. 
We minimize the distance between multi-modal context embedding and positive path with binary cross-entropy loss.


Modeling as classification, we compute the path probability score $s \in \mathbb{R}^{N}$ between $z^{\prime}$ and path embedding $h_{j}^{(p)}$ with a bilinear layer $f_{bi}$,
\begin{equation}
s = p(a|Q,I,\mathcal{G}_{sub}) = f_{bi}(z^{\prime}, h_{j}^{(p)}).
\end{equation}
A binary cross-entropy loss is also applied to optimize the path probability score:
\begin{equation}\label{bce}
\mathcal{L}_{cls} = BCE(s, l_{gt}).
\end{equation}
The whole pruning and inference network can be trained end-to-end. The final loss function is
\begin{equation}
\mathcal{L} = \mathcal{L}_{cls} + \mathcal{L}_{prune}
\end{equation}

\section{Construction of \ourdataset}
\label{sec:dataset}

\begin{table}
	\caption{Statistics of ConceptNet entities in the OK-VQA dataset. ++ and + indicates the answer is an entity or relate to an entity.}
	\label{tab:stats_concept_entity}
	\resizebox{\linewidth}{!}{
	\begin{tabular}{cccccccc}
		\toprule
		\multicolumn{2}{c}{OK-VQA split}  & \multicolumn{3}{c}{ConceptNet} & \multicolumn{3}{c}{Wikipedia} \\
		\midrule
		\multirow{2}{*}{Total} & \multirow{2}{*}{15,040} & ++ & 8,684 & 57.7\% & ++  & 5,682 & 37.8\%  \\
		\multicolumn{2}{c}{}  & + & 3,266 & 79.5\% \\
		\midrule
		\multirow{2}{*}{Train} & \multirow{2}{*}{11,508} & ++ & 7,091 & 61.6\% & ++ & 4,754 & 31.6\% \\
		&  & + & 2,227 & 81.0\% \\ 
		\midrule
		\multirow{2}{*}{Test}  & \multirow{2}{*}{6,915}  & ++ & 4,548 & 65.8\% & ++ & 3,169 & 45.8\% \\
		&  & + & 1,186 & 82.9\% \\
		\bottomrule
	\end{tabular}
	}
\end{table}


\begin{table}[]\centering\small
	\caption{Comparison of the original OK-VQA and \ourdataset on number of questions.}
	\label{tab:stats_new_split}
	\begin{tabular}{ccc}
		\toprule   
		 split & OK-VQA & ConceptVQA  \\
		\midrule
		{Training Set} & 9,009 & 8,505  \\
		{Testing Set}     & 5,046 & 4,790 \\
		{Partial Open Test} & - & 492 \\
		{Open Test}     & - & 166 \\
		\bottomrule    
	\end{tabular}
	\vspace{-1em}
\end{table}

%

The most commonly used benchmark dataset OK-VQA~\cite{Marino:2019ug} contains 15,040 image-question pairs that are divided into training set and testing set.
Each image-question pair is annotated with five answers crowdsourced from Amazon Mechanical Turk.
These questions, varying in form and knowledge category, are featured to require outside knowledge to answer.
But these human-annotated answers are free-form and noisy, and can hardly be linked to existing KGs.
To perform supervised training for open-set KB-VQA, we expect all answers to be represented as an entity in the knowledge graph.  
Therefore, we propose \ourdataset, which is built on top of the original OK-VQA, to evaluate our proposed framework.

First, we sort out 15,040 valid answers appeared in the OK-VQA dataset and match each answer to ConceptNet and Wikipedia entities. The dataset statistics is summarized in~\cref{tab:stats_concept_entity}. 
We noticed 57.7\% out of all answers are the ConceptNet entities, while 37.8\% are Wikipedia entities.
Therefore, we proceed with ConceptNet as the oracle KG and leave the exploration on Wikipedia or Wikidata as future work.
Then, we use ConceptNet api to look up related entities of the other answers and keep the ones with confidence score above 0.8.
On top of matching table we obtained so far, we correct any wrongly paired answer-entity by hand, the results of which are presented in~\cref{tab:stats_new_split}.


After the matching step, over 80\% of all answers in OK-VQA are labeled with ConceptNet entities. 
Among all the 18,423 answers from training and testing set, 8,684 that are already entities so they remain unchanged, and another 7,348 answers are related to entities in ConceptNet, which are organized into 16,032 valid entity-answers pairs.

Second, we select the questions to keep in our \ourdataset. 
We make sure that each entity appears in ConceptNet-5.6.0 for further use of the pre-trained node embeddings.
After the annotating process, we keep the questions with at least one entity-answer and score each answer $a$ as original score to match the form of VQA \cite{Antol:2015wh}, which is calculated as the percentage of predicted answers that were proposed by at least 3 human annotators: 
$acc = \min(\frac{\# \text{human gave the answer}}{3}, 1)$.
We scale answer score to $1$ for questions with only one entity answer.

Finally, we split the filtered questions into training and testing set.
The reduced train and test sets of \ourdataset consist of 8,505 training questions and 4,790 testing questions respectively from the original OK-VQA, which contains 88.4\% of the original samples.
We identify certain questions at least one of whose answer never show up in the training set, which hereafter noted as open questions.

In summary, we present the \ourdataset of 8,505 training questions and 4,790 testing questions, and 166 open questions in the testing set.
Of all the annotated answers, we have 3,651 individual answer entities for training set and 2,677 for testing set and 4,025 different entities in total.

\section{Experiments}

\subsection{Implementation Details}
We conduct experiments on proposed benchmark \ourdataset. The schema graph is extracted from the entire ConceptNet.
Nodes in schema graphs are shuffled before training to avoid possible bias brought by alphabetic orders of nodes.
Triplet loss margin is 0.5. 
$\theta_p$ in Eq.~\ref{l_prune} is 0.3.
We use the drop out rate of 0.5 throughout all MLP layers. 
We keep the original learning rate 2e-5 for ViLBERT and use 1e-4 for the rest of the network with no warmup.
The number of paths sampled in each subgraph is \(N = 200\). 
Each path sampled via random walk is no longer than 3.
To enable effective path inference, we first train the pruning network for 40 epochs with path inference network frozen and then jointly train for 30 epochs.



\begin{figure}
	\centering
	\includegraphics[width=\linewidth]{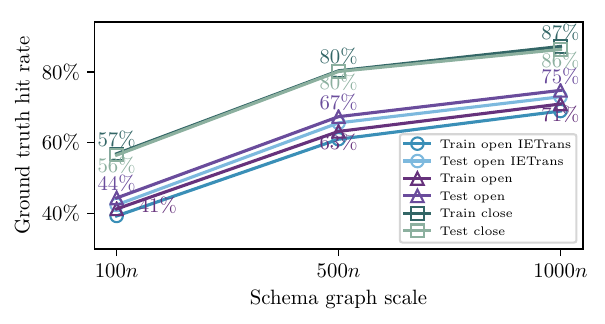}
	\caption{Hit rate of ground truths at different schema graph scales under open-set or close-set settings. The node space where schema graph pick nodes from are entire ConceptNet and candidate answers (4,025 entities in total) respectively in open and close setting. }
	\label{fig:gt_hitrate_graphsize}
\end{figure}


\begin{table}[]\centering
	\caption{Node classification recall on schema graphs of different sizes.}
	\label{tab:compare_graph_scale}
	\begin{tabular}{cccc}
		\toprule   
		& $R$@1 & $R$@10 & $R$@100 \\
		\midrule
		$100n$  & 13.39\% & 38.77\% & 52.56\% \\
		$500n$  & 11.55\% & 36.94\% & 56.70\% \\
		$1000n$ & 22.72\% & 48.33\% & 62.92\% \\
		\bottomrule    
	\end{tabular}
\end{table}

\begin{table}[]\centering
	\caption{Ground truth distribution in schema graph.
	The labels `q', `v' indicate ground truths being the key nodes extracted from question and image.  `n-1' and `n-2' represent neighbor nodes at one-hop and two-hop distance.}
	\label{tab:gt_hit_rate}
	\begin{tabular}{ccccc}
		\toprule   
        & $q$ & $v$ & $n$-1 & $n$-2 \\
		\midrule
		Training Set & 7.35\% & 11.53\% & 45.10\% & 22.89\%\\
		Testing Set & 6.85\% & 11.86\% & 48.79\% & 23.38\%\\
		\bottomrule    
	\end{tabular}	
	\vspace{-1em}
\end{table}

\begin{table*}[]\centering\small
  \caption{Ablation studies on \ourdataset. `w/o Prune' and `Node Cls' indicate skipping pruning and path sampling. The performance is measured in top k Recall (\%). }
  \label{tab:ablation}
  \setlength{\tabcolsep}{2mm}{
  \begin{tabular}{cccccccccc}
  \toprule
  	\multirow{2}{*}{Pruning} & \multirow{2}{*}{Inference} & \multirow{2}{*}{Ablation} & \multicolumn{2}{c}{Top 1 Recall}& \multicolumn{3}{c}{Rank by Node} & \multicolumn{2}{c}{Rank by Path} \\
  	\cmidrule(lr){4-5} \cmidrule(lr){6-8} \cmidrule(lr){9-10} 
  	&&& full test & partial open & $R$@10 & $R$@50 & $R$@100  & $R$@10 & $R$@20 \\
  	\midrule
  	\multirow{2}{*}{$\mathcal{L}_{prune}$ only} & Node Cls & all triplets & 11.16 & 2.97 & 39.25 & 57.43 & 62.27 & - & - \\
  	 & Node Cls & semi-hard & 22.72 & 5.93 & 48.33 & 59.36 & 62.92 & - & - \\
  	\midrule
  	\multirow{5}{2cm}{$\mathcal{L}_{cls} + \mathcal{L}_{prune}$ (prune to $100$)} 
  	& Path & all triplets & 18.86 & 5.98 & 45.44 & 52.10 & 59.37 & 37.78 & 44.61 \\
	& Path & w/o ptm \& node-type & 24.11 & 7.44 & 47.77 & 59.42 & 63.10 & 41.24 & 47.53\\
  	& Path & 1-layer Trans & 17.47 & 5.53 & 41.77 & 55.02 & 63.17 & 37.65 & 44.14\\
  	& Path & LSTM & 24.68 & 6.71 & 47.14 & 59.21 & 63.06 & 41.33 & 46.20\\
  	& Path & \ourmethod(ours) & 24.94 & 7.60 & 49.47 & 60.45 & 63.07 & 42.26 & 48.71 \\
  	\bottomrule  
  \end{tabular}}
\end{table*}

\subsection{Ablation Studies}
\label{subsec:ablation}

\textbf{Schema Graph Quality.}
To verify that the retrieved schema graph can cover the ground-truth answers as complete as possible, we investigate the hit rate of ground truths at different graph scales.
As in Fig.~\ref{fig:gt_hitrate_graphsize}, a satisfiable total of 72.3\% schema graphs retrieve ground truth answers within 1,000 nodes and 44.3\% in 100 nodes, proving schema graphs construction functioning at the sacrifice of a certain number of ground truth answers.
Under the same ground truth scoring protocol of VQA\cite{Antol:2015wh}, the score of retrieved answers on testing set is 68.80\%, which is also the upper-bound of experimental results below. 

In addition, we explore the distribution of distance from our defined key nodes to ground truth entities.
In Tab.~\ref{tab:gt_hit_rate}, nearly all the answers can be retrieved within two steps, which again confirms that our schema graph construction is reasonable.

\textbf{Sensitivity to SGG.} On account of scene graph generation methods that takes a role in schema graph construction, we perform ablative experiments to evaluate framework sensitivity to scene graph generation methods.
We compare the schema graphs constructed with two recent scene graph generation models, Causal-TDE~\cite{Tang:2020sg} (used in our framework) and IETrans~\cite{zhang2022fine}.
The hit rate of ground truth nodes are on par with our original results, as depicted in Fig.~\ref{fig:gt_hitrate_graphsize}, proving robustness of our framework to scene graph generation methods.

\textbf{Network Pruning.}
To demonstrate the effectiveness of graph pruning module, we perform multi-label classification on schema graphs of various scales.
As shown in Tab.~\ref{tab:compare_graph_scale}, node classification on 1000-node schema graphs achieves 62.92 $R$@100, which is higher than 52.56 $R$@100 of 100-node schema graphs.
We conclude that the size of 1,000 is suitable for our framework, which results in highest recall at top 100.
This indicates that the graph pruning stage can effectively remove noisy node from the schema graph and prepare a feasible scale of candidates for path sampling.

\begin{figure}[t]
  \vspace{-1em}
  \centering
  \includegraphics[width=\linewidth]{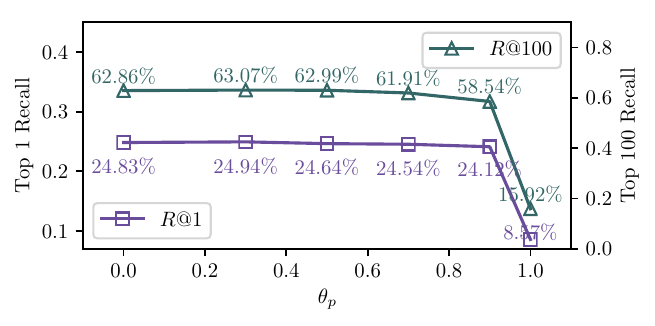}
  \caption{Sensitivity analysis of $\theta_p$. $R$@100 results of our framework under different $\theta_p$.}
  \vspace{-1em}
  \label{fig:sensitivity-theta}
\end{figure}

We conduct sensitivity analysis of the pruning hyper-parameter $\theta_p$ in Eq.~\ref{l_prune}, which controls connectivity of the pruned graph.
Higher $\theta_p$ guarantees better connectivity.
As shown in Fig.~\ref{fig:sensitivity-theta}, our results of $R$@1 and $R$@100 is insensitive to $\theta_p$ within a reasonable range, achieving adaptive balance between node semantics and graph connectivity. 

\textbf{Modular Ablation.}
We perform modular ablation analysis in Tab.~\ref{tab:ablation}.
The upper half of the table indicates results of node classification on the 1,000n schema graph, with $\mathcal{L}_{prune}$ supervising node encoding. The node-level predictions are obtained by computing cosine similarity between node features and multimodal query embedding. 
The lower half contains results of path ranking on graphs pruned to 100-node. $\mathcal{L}_{prune}$ supervises node encoding for pruning while $\mathcal{L}_{cls}$ supervises path ranking.

First we evaluate the semi-hard triplet loss adopted for computing $\mathcal{L}_{prune}$.
Comparing to using all triplets, mining semi-hard triplets improves node classification accuracy and speed by a large margin ($R$@1 doubled).
Semi-hard triplets subsequently affects path ranking as well. \ourmethod achieves a 6\% gain in $R$@1 than trained with all triplets.
Next we investigate the effectiveness of node encoding (Eq.~\ref{node_enc}). We observe a slight decrease in node and path recalls once we remove ptm and node-type encoding, indicating that it is helpful in learning node features. 

Then we evaluate path encoder for information aggregation ability (\cref{p_encode}) among several choices: 1-layer transformer, LSTM, and 2-layer MLP~(ours).
The one-layer transformer is proved inadequate for our path encoding since the head and tail of each path should be treated differently.
Although LSTM and classical MLP can both capture sequential relations, MLP outperforms LSTM in terms of path recall and inference speed.
Given the relatively short inference paths (within 3 steps), the fully-connected nature of MLP can best encode our inference paths without the need of learning long-term dependencies.

Additionally, we present result of our ablated model with a classifier \ourmethod + w/ cls in \cref{tab:baseline-cls}, to verify that schema graph construction can effectively distill knowledge. 
We input the schema graph obtained after \cref{subsec:sgc} and \cref{subsec:prune} into a GAT layer. With the same query encoder ViLBERT, we then predict answer distribution with the concatenation of encoded query and pooled graph representation.


\subsection{Comparison to SOTA Approaches}
\label{subsec:mainexp}

\begin{table}[t]\centering\small
  \caption{Accuracy (\%) of classification baselines under close-set setting, indicating limited answer space of training and testing gt answers.  $R$@10 reported in `open' column.}
  \label{tab:baseline-cls}
  \setlength{\tabcolsep}{1mm}{
  \begin{tabular}{ccccc}
  \toprule
  	 Models & Answer Space  & test & partial open & open\\
  	\midrule
  	LXMERT  & 4,025  & 21.92  & 5.16 & 0.00 \\
  	ViLBERT & 4,025  & 38.78 & 9.84 & 0.00 \\
  	MAVEx(ViL) \cite{Wu:2021ma}  & 5,020  & 35.20 & 10.69 & 0.60\\
  	MAVEx \cite{Wu:2021ma}   & 5,020  & 39.40 & 8.74 & 0.60\\
  	\midrule
  	\ourmethod + w/ cls & 5,020 & 38.24 & 11.02 & 0.72 \\
  	\midrule
  	\ourmethod (ours) & 516,782 & 24.94* & 7.60* & 3.01*\\
  	\bottomrule
  \end{tabular}}
  \vspace{-1em}
\end{table}


Compared to our proposed open-set framework, we evaluate current SOTA methods under close-set setting on \ourdataset and OK-VQA in \cref{tab:baseline-cls}.
The close-set answer spaces of \ourdataset and OK-VQA are 4,025 and 5,020 respectively, which contain all candidates from train and test sets.
Our results (*) in the last row are obtained under open-set setting, which allows 100-fold increase in answer space.

As described in \cref{sec:dataset}, to investigate model performance on truly open-ended questions, we identifies 492 partially-open question and 166 open questions whose answer never appear in the training data. 
Since answers to these questions are unseen during training, we report the $R$@10 on open questions which are extremely hard.
Retriever-classifier methods LXMERT\cite{Tan:2019lx}, ViLBERT\cite{Lu:2019vilbert}, and MAVEx\cite{Wu:2021ma} fail to solve open questions, while our framework shows less discrimination between partial-open and open questions.
When we extend the answer space to 500k, both LXMERT and ViLBERT result in out-of-memory. 

For fair comparison under close-set setting, we present \ourmethod with classification on OK-VQA, which is marked as \ourmethod + w/ cls. 
The accuracy 38.24\% is higher than baseline 35.20\% reported in \cite{Wu:2021ma}, proving that our schema graph construction process can focus on the informative node in KGs and surpass state-of-the-art methods based on classification.


Additionally, recent works~\cite{Tsimpoukelli:2021mf, Jin:2022AGP, Yang:2022ae} show that large scale generative methods is able to answer zero-shot or few-shot KBVQA questions. However, these methods rely on large-scaled training corpus which limits their applicability and reliability. In comparison, our method digs into the external knowledge and provides answers with path-level rationale. We will further provide in-depth comparison with LLMs in Appendix E.

\subsection{Explainable Path Reasoning}
\begin{figure}[t]
  \centering
  \includegraphics[width=\linewidth]{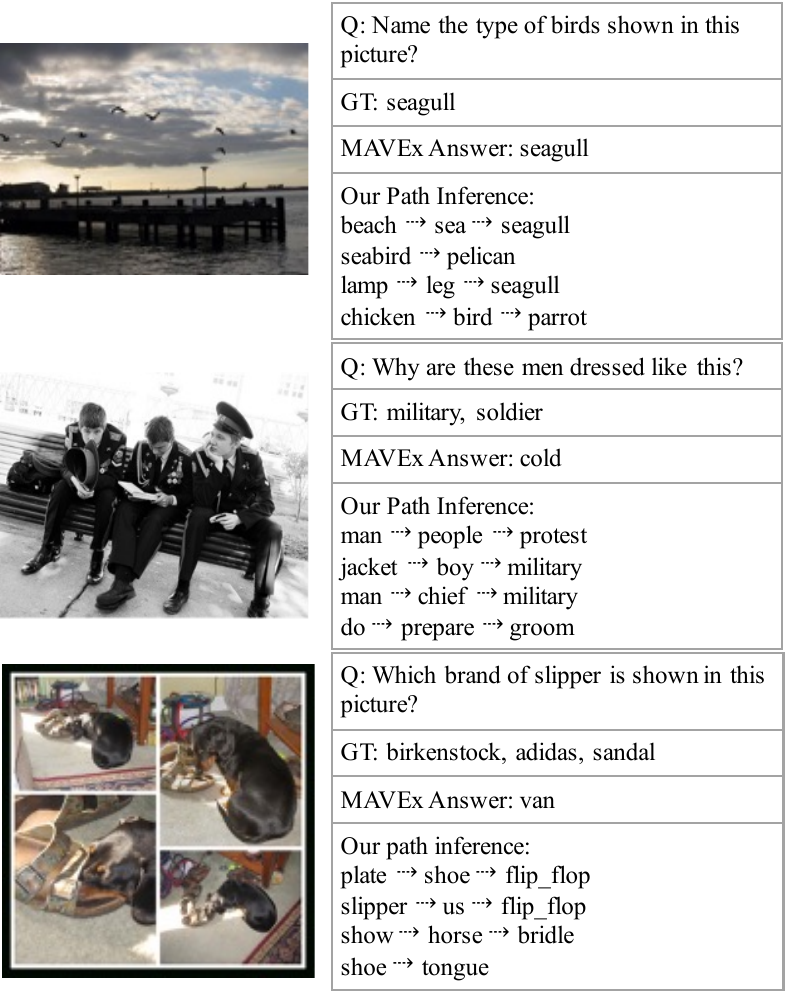}
  \caption{Examples of path inference. The first example is answered correctly. The last example is an open question. Paths are presented in ranked order.}
  \vspace{-1em}
  \label{fig:zero-shot-example}
\end{figure}

Another important goal of \ourmethod is to provide explainable reasoning path accompanied with the predicted answers. It is an attempt to apply structured knowledge explicitly in KGs.
\cref{fig:zero-shot-example} provides visualization of inference paths predicted from \ourmethod.
The first question is a common example of VQA questions. 
The answer `seagull' can be reasoned from abundant information and retrieved knowledge: the indicator word `birds', the visually recognized object `lamp' and the associated entities `beach', `sea', `seabird'.
The second question is a more typical KB-VQA question.
In this example, visual clues play are vitae in reasoning. `jacket' and `man' could associate with scenarios like `protest', `military' and 'groom', which may not be the correct answer but are plausible guesses. 

The last example in \cref{fig:zero-shot-example} is a zero-shot question.
GT answers `birkenstock', `adidas' and `sandal' are not seen during training. 
Compared to classification baseline that produces an irrelevant guess `van', ours locates on a close answer `flip-flop', showing its generalization ability on zero-shot questions.
Our reasoning are performed from the queried key works `slipper' and `brand'.


\section{Conclusion}


The large number of nodes and structured relationships make KB-VQA more challenging compared with traditional VQA. 
Most of existing works on KB-VQA ignore these properties of KGs but only optimize the likelihood of the ground-truth answers among a small range of human-annotated candidates. It actually have no fundamental difference from the traditional VQA.
This paper presents a new Open-set KB-VQA as well as a corresponding benchmark, \ourdataset, where the answer space is extended to any entity in the knowledge graph. 
Starting from this problem, we propose a new retriever-ranker paradigm \ourmethod, which is able to not only produces answers across the entire KG but also acquire extensibility in terms of knowledge-centric commonsense reasoning. 
\ourmethod demonstrate the possibility of path-level inference directly on large-scale knowledge graphs. 

Still, it is a long way to go for realizing convincing inference path reasoning. 
From the dataset perspective, \ourdataset simply regards any path ended with the ground-truth as positive. 
It requires accurate reasoning path annotations apart from the ground-truth answers for both training and evaluation. 
From the method perspective, our path sampling strategy is relatively rudimentary. 
The relationship between the question logic and the reasoning path is not fully utilized.
We hope our work can inspire further researches on open-set KB-VQA, improving applicability of VQA in real world settings.

\clearpage
{\small
\bibliographystyle{ieee_fullname}
\bibliography{reference}

\begin{thebibliography}{10}\itemsep=-1pt

\bibitem{Antol:2015wh}
Stanislaw Antol, Aishwarya Agrawal, Jiasen Lu, Margaret Mitchell, Dhruv Batra,
  C~Lawrence Zitnick, and Devi Parikh.
\newblock {VQA: Visual Question Answering.}
\newblock {\em ICCV}, 2015.

\bibitem{Auer:2007db}
S. Auer, Christian Bizer, Georgi Kobilarov, Jens Lehmann, Richard Cyganiak, and
  Zachary~G. Ives.
\newblock Dbpedia: A nucleus for a web of open data.
\newblock In {\em ISWC/ASWC}, 2007.

\bibitem{Brown:2020gpt3}
Tom Brown, Benjamin Mann, Nick Ryder, Melanie Subbiah, Jared~D Kaplan, Prafulla
  Dhariwal, Arvind Neelakantan, Pranav Shyam, Girish Sastry, Amanda Askell,
  Sandhini Agarwal, Ariel Herbert-Voss, Gretchen Krueger, Tom Henighan, Rewon
  Child, Aditya Ramesh, Daniel Ziegler, Jeffrey Wu, Clemens Winter, Chris
  Hesse, Mark Chen, Eric Sigler, Mateusz Litwin, Scott Gray, Benjamin Chess,
  Jack Clark, Christopher Berner, Sam McCandlish, Alec Radford, Ilya Sutskever,
  and Dario Amodei.
\newblock Language models are few-shot learners.
\newblock In H. Larochelle, M. Ranzato, R. Hadsell, M.F. Balcan, and H. Lin,
  editors, {\em Advances in Neural Information Processing Systems}, volume~33,
  pages 1877--1901. Curran Associates, Inc., 2020.

\bibitem{Ding:2022vt}
Yang Ding, Jing Yu, Bang Liu, Yue Hu, Mingxin Cui, and Qi Wu.
\newblock {MuKEA: Multimodal Knowledge Extraction and Accumulation for
  Knowledge-based Visual Question Answering}.
\newblock {\em arXiv.org}, Mar. 2022.

\bibitem{Feng:2020tl}
Yanlin Feng, Xinyue Chen, Bill~Yuchen Lin, Peifeng Wang, Jun~Yan 0012, and
  Xiang~Ren 0001.
\newblock {Scalable Multi-Hop Relational Reasoning for Knowledge-Aware Question
  Answering.}
\newblock {\em EMNLP}, 2020.

\bibitem{Gao:2022gw}
Feng Gao, Qing Ping, Govind Thattai, Aishwarya Reganti, Ying~Nian Wu, and Prem
  Natarajan.
\newblock {A Thousand Words Are Worth More Than a Picture: Natural
  Language-Centric Outside-Knowledge Visual Question Answering}.
\newblock Jan. 2022.

\bibitem{GarciaOlano:2022ia}
Diego Garcia-Olano, Yasumasa Onoe, and Joydeep Ghosh.
\newblock Improving and diagnosing knowledge-based visual question answering
  via entity enhanced knowledge injection.
\newblock {\em Companion Proceedings of the Web Conference 2022}, 2022.

\bibitem{Garderes:2020tc}
Fran{\c c}ois Gard{\`e}res, Maryam Ziaeefard, Baptiste Abeloos, and Freddy
  L{\'e}cu{\'e}.
\newblock {ConceptBert - Concept-Aware Representation for Visual Question
  Answering.}
\newblock {\em EMNLP}, 2020.

\bibitem{Gui:2022kat}
Liangke Gui, Borui Wang, Qiuyuan Huang, Alexander~G. Hauptmann, Yonatan Bisk,
  and Jianfeng Gao.
\newblock Kat: A knowledge augmented transformer for vision-and-language.
\newblock In {\em NAACL}, 2022.

\bibitem{He:2017mrcnn}
Kaiming He, Georgia Gkioxari, Piotr Dollár, and Ross Girshick.
\newblock Mask r-cnn.
\newblock In {\em 2017 IEEE International Conference on Computer Vision
  (ICCV)}, pages 2980--2988, 2017.

\bibitem{Jiang:2022jj}
Jinhao Jiang, Kun Zhou, Wayne~Xin Zhao, and Ji-Rong Wen.
\newblock {Great Truths are Always Simple: A Rather Simple Knowledge Encoder
  for Enhancing the Commonsense Reasoning Capacity of Pre-Trained Models}.
\newblock May 2022.

\bibitem{Jin:2022AGP}
Woojeong Jin, Yu Cheng, Yelong Shen, Weizhu Chen, and Xiang Ren.
\newblock A good prompt is worth millions of parameters: Low-resource
  prompt-based learning for vision-language models.
\newblock {\em ArXiv}, abs/2110.08484, 2022.

\bibitem{Karpukhin:2020dpr}
Vladimir Karpukhin, Barlas Oğuz, Sewon Min, Patrick Lewis, Ledell~Yu Wu,
  Sergey Edunov, Danqi Chen, and Wen tau Yih.
\newblock Dense passage retrieval for open-domain question answering.
\newblock {\em ArXiv}, abs/2004.04906, 2020.

\bibitem{li2019visualbert}
Liunian~Harold Li, Mark Yatskar, Da Yin, Cho-Jui Hsieh, and Kai-Wei Chang.
\newblock Visualbert: A simple and performant baseline for vision and language.
\newblock {\em arXiv preprint arXiv:1908.03557}, 2019.

\bibitem{Liu:2004cn}
Hugo Liu and Push Singh.
\newblock Conceptnet — a practical commonsense reasoning tool-kit.
\newblock {\em BT Technology Journal}, 22:211--226, 2004.

\bibitem{Lu:2019vilbert}
Jiasen Lu, Dhruv Batra, Devi Parikh, and Stefan Lee.
\newblock Vilbert: Pretraining task-agnostic visiolinguistic representations
  for vision-and-language tasks.
\newblock In {\em NeurIPS}, 2019.

\bibitem{Luo:2021up}
Man Luo, Yankai Zeng, Pratyay Banerjee, and Chitta Baral.
\newblock {Weakly-Supervised Visual-Retriever-Reader for Knowledge-based
  Question Answering.}
\newblock {\em EMNLP}, 2021.

\bibitem{Marino:2021tn}
Kenneth Marino, Xinlei Chen, Devi Parikh, Abhinav~Gupta 0001, and Marcus
  Rohrbach.
\newblock {KRISP - Integrating Implicit and Symbolic Knowledge for Open-Domain
  Knowledge-Based VQA.}
\newblock {\em CVPR}, 2021.

\bibitem{Marino:2019ug}
Kenneth Marino, Mohammad Rastegari, Ali Farhadi, and Roozbeh Mottaghi.
\newblock {OK-VQA - A Visual Question Answering Benchmark Requiring External
  Knowledge.}
\newblock {\em CVPR}, 2019.

\bibitem{Narasimhan:2018ot}
Medhini Narasimhan, Svetlana Lazebnik, and Alexander~G. Schwing.
\newblock Out of the box: Reasoning with graph convolution nets for factual
  visual question answering.
\newblock In {\em NeurIPS}, 2018.

\bibitem{Shevchenko:2021ro}
Violetta Shevchenko, Damien Teney, Anthony~R. Dick, and Anton van~den Hengel.
\newblock Reasoning over vision and language: Exploring the benefits of
  supplemental knowledge.
\newblock {\em ArXiv}, abs/2101.06013, 2021.

\bibitem{Tan:2019lx}
Hao Tan and Mohit Bansal.
\newblock Lxmert: Learning cross-modality encoder representations from
  transformers.
\newblock {\em ArXiv}, abs/1908.07490, 2019.

\bibitem{Tang:2020sg}
Kaihua Tang, Yulei Niu, Jianqiang Huang, Jiaxin Shi, and Hanwang Zhang.
\newblock Unbiased scene graph generation from biased training.
\newblock {\em 2020 IEEE/CVF Conference on Computer Vision and Pattern
  Recognition (CVPR)}, pages 3713--3722, 2020.

\bibitem{Tsimpoukelli:2021mf}
Maria Tsimpoukelli, Jacob Menick, Serkan Cabi, S.~M.~Ali Eslami, Oriol Vinyals,
  and Felix Hill.
\newblock Multimodal few-shot learning with frozen language models.
\newblock In {\em NeurIPS}, 2021.

\bibitem{Wang:2022wi}
Kuan Wang, Yuyu Zhang, Diyi Yang, Le Song, and Tao Qin.
\newblock {GNN is a Counter? Revisiting GNN for Question Answering.}
\newblock {\em ICLR}, 2022.

\bibitem{Wang:2022bo}
Yanan Wang, Michihiro Yasunaga, Hongyu Ren, Shinya Wada, and Jure Leskovec.
\newblock Vqa-gnn: Reasoning with multimodal semantic graph for visual question
  answering.
\newblock {\em ArXiv}, abs/2205.11501, 2022.

\bibitem{wang2021simvlm}
Zirui Wang, Jiahui Yu, Adams~Wei Yu, Zihang Dai, Yulia Tsvetkov, and Yuan Cao.
\newblock Simvlm: Simple visual language model pretraining with weak
  supervision.
\newblock {\em ICLR}, 2021.

\bibitem{Wu:2021ma}
Jialin Wu, Jiasen Lu, Ashish Sabharwal, and Roozbeh Mottaghi.
\newblock {Multi-Modal Answer Validation for Knowledge-Based VQA.}
\newblock {\em AAAI}, 2103:arXiv:2103.12248, 2021.

\bibitem{Yang:2021jb}
Zhengyuan Yang, Zhe Gan, Jianfeng Wang, Xiaowei Hu, Yumao Lu, Zicheng Liu, and
  Lijuan Wang.
\newblock {An Empirical Study of GPT-3 for Few-Shot Knowledge-Based VQA}.
\newblock Sept. 2021.

\bibitem{Yang:2022ae}
Zhengyuan Yang, Zhe Gan, Jianfeng Wang, Xiaowei Hu, Yumao Lu, Zicheng Liu, and
  Lijuan Wang.
\newblock An empirical study of gpt-3 for few-shot knowledge-based vqa.
\newblock In {\em AAAI}, 2022.

\bibitem{Yasunaga:2021ug}
Michihiro Yasunaga, Hongyu Ren, Antoine Bosselut, Percy Liang, and Jure
  Leskovec.
\newblock {QA-GNN - Reasoning with Language Models and Knowledge Graphs for
  Question Answering.}
\newblock {\em NAACL-HLT}, 2021.

\bibitem{Lin:2019vf}
Bill Yuchen~Lin, Xinyue Chen, Jamin Chen, and Xiang Ren.
\newblock {KagNet: Knowledge-Aware Graph Networks for Commonsense Reasoning}.
\newblock {\em CoRR}, page arXiv:1909.02151.

\bibitem{zeng2022socratic}
Andy Zeng, Adrian Wong, Stefan Welker, Krzysztof Choromanski, Federico Tombari,
  Aveek Purohit, Michael Ryoo, Vikas Sindhwani, Johnny Lee, Vincent Vanhoucke,
  et~al.
\newblock Socratic models: Composing zero-shot multimodal reasoning with
  language.
\newblock {\em arXiv preprint arXiv:2204.00598}, 2022.

\bibitem{zhang2022fine}
Ao Zhang, Yuan Yao, Qianyu Chen, Wei Ji, Zhiyuan Liu, Maosong Sun, and Tat-Seng
  Chua.
\newblock Fine-grained scene graph generation with data transfer.
\newblock In {\em ECCV}, 2022.

\end{thebibliography}
}

\clearpage

\section*{Appendices}

\begin{table*}[t]\centering\small
  \caption{Effect of answer space and zero-shot questions.}
  \label{tab:open-results}
  \begin{tabular}{ccccccccc}
  \toprule
  	  	\multirow{2}{*}{open / close} & \multirow{2}{*}{ablation} & \multicolumn{2}{c}{Top 1 Recall}& \multicolumn{3}{c}{Rank by Node} & \multicolumn{2}{c}{Rank by Path} \\
  	\cmidrule(lr){3-4} \cmidrule(lr){5-7} \cmidrule(lr){8-9} 
  	&& full test & partial open & $R$@10 & $R$@50 & $R$@100  & $R$@10 & $R$@20 \\
  	\midrule
  	open & \ourmethod & 24.94 & 7.60 & 49.47 & 60.45 & 63.07 & 42.26 & 48.71 \\
  	\midrule
  	\multirow{2}{*}{close}&  $\mathcal{L}_{prune}$ only& 26.26 & 7.23 & 55.88 &  68.37 & 71.60 & - & -\\
  	& \ourmethod & 28.14 & 6.63 & 56.55 & 62.18 & 62.18 & 40.81 & 48.18 \\
  	\bottomrule  
  \end{tabular}
\end{table*}

\begin{figure*}[t]
  \centering
  \includegraphics[width=\linewidth]{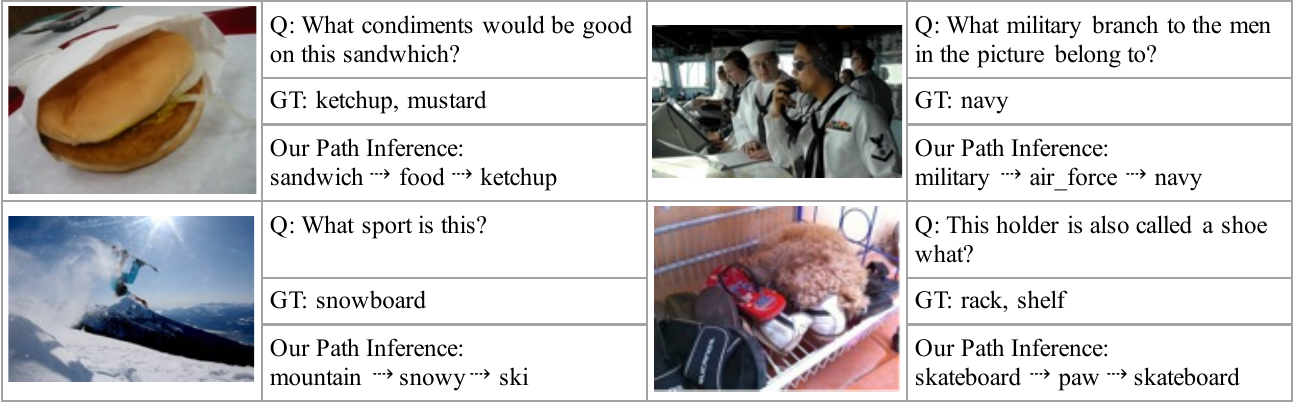}
  \caption{More examples of our inference results. Top row are correct predictions and bottom row are false predictions.}
  \vspace{-1em}
  \label{fig:more_examples}
\end{figure*}

\section*{A. Analysis on Close-set Testing}


As mentioned in \cref{subsec:ablation}, our proposed framework has the ability of predicting open-set answers from all 516,782 ConceptNet entities, while most counterparts perform classification on close-set answer candidates. 
To make a fairer comparison with existing classification baselines, we examine our proposed \ourmethod on close-set answer candidates as the extended analysis of results in \cref{tab:baseline-cls}.
%
We propose a close-set setting that adapts the schema graph constructing process to recruit entities that appear in the 4,025 answer candidates only.
While building close-set schema graph, we still keep the visual and textual entities as key nodes, but only candidate entities can be retrieved as $\mathcal{V}_N$ in the schema graph. 
A similar path sampling rule is also applied.
As portrayed in \cref{fig:gt_hitrate_graphsize}, the ground truth hit rate of close-set schema graph achieves 80\% at graph size of 500, and 87\% at 1,000.
To keep consistency with the hit rate of 72.5\% for 1,000-node open-set schema graph, we report performance on close-set schema graphs of size 500. 

In \cref{tab:open-results}, we perform open-set and close-set experiments on \ourdataset.
The close-set schema graph boost the accuracy on R@1, indicating narrowing down answer space would simplify the task.
Ablation experiments with close-set schema graphs result in similar patterns that path sampling and semi-hard triplet loss would increase accuracy.
The close-set models could achieve better performance if tested with a more specifically designed sampling module.

In comparison, we performed an open-set experiment on two baseline methods LXMERT and VilBERT to simulate the open-set setting.
Both methods uses all ground truth answers appeared in training and testing data as the answer space which is 4,025 answers in our constructed \ourdataset.
But the classifier predict correct answer for none of the open questions, as reported in \cref{tab:baseline-cls}, .
To fill the gap in answer space size between the answer candidate and ConceptNet we use, we extend their answer space to a shuffled 50,000-class and then retrain the classifier. 
LXMERT and VilBERT perform identically to their open-set results, which indicates that both classifiers learn biased answer distribution and are unable to answer open questions.

\section*{B. Schema Graph Construction Details}
We defined 21 different relations in schema graph construction.
The relations are \textit{
    'antonym',
    'atlocation',
    'capableof',
    'causes',
    'createdby',
    'derivedfrom',
    'desires',
    'hasa',
    'hascontext',
    'hasproperty',
    'hassubevent',
    'isa',
    'madeof',
    'mannerof',
    'notcapableof',
    'notdesires',
    'partof',
    'receivesaction',
    'relatedto',
    'synonym',
    'usedfor'}.
    
We specify the node ranking process mentioned in schema graph construction.
It takes 3 steps to rank the retrieved neighbor nodes.

\textit{Step 1.} We exclude nodes that do not connect to any key nodes.

\textit{Step 2.} We rank the nodes by its relation to the key nodes.
We counted the relations that connects the ground truth node to key nodes and ranked the relation types accordingly.

\section*{C. Comparison to Generative Methods.}

In \cref{tab:baseline-gen}, we present performance of our \ourmethod tested on \ourdataset along with other generative baselines for Open-set KB-VQA. 
Answer space of any generative method is the vocabulary size of the dependent language model, which is fixed unless re-training the entire backbone model.
Instead, our framework allows a bigger and more flexible answer space that varies with the size of the knowledge graph.

Generative models equipped with knowledge from multiple KGs can perform truly open-ended VQA.
These models that trained on several VQA datasets excluding OK-VQA, achieve great zero-shot performance on the original test set of 5.9\%, 11.6\% and 17.5\% respectively, which can be considered open-set here. 
We evaluate supervised PICa and our method on the partial open set of \ourdataset. 
In Open-set test, we takes the 516,782 entities in the entire ConceptNet as answer candidates.

\section*{D. Qualitative Analysis}

We show more inference examples in \cref{fig:more_examples}.
The two examples in the top row are answered correctly.
For the example at bottom left, the predicted answer is wrong but the inference path is reasonable.
The correctly spotted visual clues 'mountain' and 'snowy' could lead to 'ski' or 'snowboard'.
However, the mis-prediction indicates that our framework need better ability of distinguishing visual details e.g. actions.
Our framework fail to predict answer for the bottom right example because the ground truth entities 'rack' and 'shelf' are not included in the schema graph.
As stated in \cref{fig:gt_hitrate_graphsize}, the ground truth of 72.5\% examples are retrieved through our proposed schema graph construction, which leaves us room for improvement.

\begin{table}[H]\centering\small
  \caption{Open-set results of generative baselines. Open-set means model answers with open answers from KGs or large vocabulary bank.\vspace{1ex}}
  \label{tab:baseline-gen}
  \resizebox{\linewidth}{!}{
  \begin{tabular}{cccc}
  \toprule
  	 Models & Answer Space & Zero-shot & Partial open* \\
  	\midrule
  	Frozen \cite{Tsimpoukelli:2021mf} & Vocab~(32k) & 5.9 \\
  	$\text{FewVLM}_{base}$ \cite{Jin:2022AGP} & GPT-3 Vocab~(50k) & 11.6 & \\
  	$\text{PICa}_{full}$ \cite{Yang:2022ae} & GPT-3 Vocab~(50k) & 17.5 & 30.53 \\
  	\midrule
  	\ourmethod (ours) & ConceptNet~(516,782) & - & 7.60\\
  	\bottomrule
  \end{tabular}}
\end{table}

\end{document}